\documentclass[10pt, conference, compsocconf]{IEEEtran}
\usepackage{graphicx}
\usepackage{todonotes}
\usepackage{hyperref}
\usepackage{xspace}
\usepackage{amsmath,amsthm}

\newcommand{\dlr}{$D_\textrm{\textit{LR}}$\xspace}
\newcommand{\dhr}{$D_\textrm{\textit{HR}}$\xspace}
\newcommand{\lmdlr}{$\textrm{\textit{LM}}(D_\textrm{\textit{LR}})$\xspace}
\newcommand{\lmdhr}{$\textrm{\textit{LM}}(D_\textrm{\textit{HR}})$\xspace}
\newcommand{\svrdhr}{$\textrm{\textit{SVR}}(D_\textrm{\textit{HR}})$\xspace}
\newcommand{\svrdlr}{$\textrm{\textit{SVR}}(D_\textrm{\textit{LR}})$\xspace}
\begin{document}

\title{Automatic Extraction of Personality from Text:\\Challenges and Opportunities}

\author{
\IEEEauthorblockN{Nazar Akrami}
\IEEEauthorblockA{Department of Psychology\\
Uppsala University\\
Uppsala, Sweden\\
Email: nazar.akrami@psyk.uu.se}
\and
\IEEEauthorblockN{Johan Fernquist, Tim Isbister, Lisa Kaati, Bj\"orn Pelzer}
\IEEEauthorblockA{Decision Support Systems\\
Swedish Defence Research Agency\\
Kista, Sweden\\
Email: firstname.lastname@foi.se}
}

\maketitle

\begin{abstract}
In this study, we examined the possibility to extract personality traits from a text. We created an extensive dataset by having experts annotate personality traits in a large number of texts from multiple online sources. From these annotated texts, we selected a sample and made further annotations ending up in a large low-reliability dataset and a small high-reliability dataset. We then used the two datasets to train and test several machine learning models to extract personality from text, including a language model. Finally, we evaluated our best models in the wild, on datasets from different domains. Our results show that the models based on the small high-reliability dataset performed better (in terms of $\textrm{R}^2$) than models based on large low-reliability dataset. Also, language model based on small high-reliability dataset performed better than the random baseline. Finally, and more importantly, the results showed our best model did not perform better than the random baseline when tested in the wild. Taken together, our results show that determining personality traits from a text remains a challenge and that no firm conclusions can be made on model performance before testing in the wild.

\end{abstract}


\IEEEpeerreviewmaketitle

\section{Introduction}\label{sec:introduction}
Since the introduction of the personality concept, psychologists have worked to formulate theories and create models describing human personality and reliable measure to accordingly. The filed has been successful to bring forth a number of robust models with corresponding measures. One of the most widely accepted and used is the Five Factor Model \cite{mccrae1999five}. The model describes human personality by five traits/factors, popularly referred to as the \emph{Big Five} or OCEAN: \underline{O}penness to experience, \underline{C}onscientiousness, \underline{E}xtraversion, \underline{A}greeableness, and emotional stability\footnote{%
Normally named \emph{\underline{N}euroticism},
but we will use the inverse \emph{stability} throughout the paper to ensure consistently ``positive'' factor names, giving us a uniform correspondence between negative numerical factor values and negative personalities vs. positive values and positive personalities.} (henceforth Stability). There is now an extensive body of research showing that these factors matter in a large number of domains of people’s life. Specifically, the Big Five factors have been found to predict life outcomes such as health, longevity, work performance, interpersonal relations, migration and social attitudes, just to mention some domains (e.g. \cite{akrami2011generalized, Jokela725, Kristof2005, roberts2007power}).
To date, the most common assessment of personality is by self-report questionnaires \cite{kagan2007trio}.

In the past decade however, personality psychologist, together with computer scientist, have worked hard to solve the puzzle of extracting a personality profile (e.g., the Big Five factors) of an individual based on a combination of social media activities \cite{Schwartz2013}. However, in the aftermath of Cambridge Analytica scandal, where the privacy of millions of Facebook users was violated, this line of research has been met with skepticism and suspicion. More recent research focuses on text from a variety of sources, including twitter data (e.g. \cite{arnoux201725, majumder2017}). Recent development in text analysis, machine learning, and natural language models, have move the field into an era of optimism, like never before. Importantly, the basic idea in this research is that personality is reflected in the way people write and that written communication includes information about the author’s personality characteristics \cite{PennebakerKing1999}. Nevertheless, while a number of attempts has been made to extract personality from text (see below), the research is standing remarkably far from reality. There are, to our knowledge, very few attempts to test machine learning models ``in the wild''. The present paper aims to deal with this concern. Specifically, we aim to (\textbf{A}) create a model which is able to extract Big Five personality from a text using machine learning techniques, (\textbf{B}) investigate whether a model trained on a large amount of solo-annotated data performs better than a model trained on a smaller amount of high quality data, and, (\textbf{C}) measure the performance of our models on data from another two domains that differ from the training data.

\section{Related Work}\label{sec:related_work}
In \cite{Kosinski2013} the authors trained a combination of logistic and linear regression models on data from 58,466 volunteers, including their demographic profiles, Facebook data and psychometric test results, such as their Big Five traits.
This data, the \emph{myPersonality} dataset \cite{myPersonality}, was available for academic research until 2018,
although this access has since been closed down.
A demonstration version of the trained system is available to the public in form of the \emph{ApplyMagicSauce} web application\footnote{\url{https://applymagicsauce.com}} of Cambridge University.

In 2018 media exposed the unrelated (and now defunct) company \emph{Cambridge Analytica} to considerable public attention for having violated the privacy and data of millions of Facebook users and for having meddled in elections, with some of these operations misusing the aforementioned research results.
This scandal demonstrates the commercial and political interest in this type of research,
and it also emphasizes that the field has significant ethical aspects.

Several attempts have been made to automatically determining the Big Five personality traits using only text written by the test person.
A common simplification in such
approaches is to model each trait as binary (\emph{high} or \emph{low}) rather than on a more
realistic granular spectrum.

The authors of \cite{argamon2009automatically} trained a Bayesian Multinomial Regression model on stylistic and content features of a collection of student-written stream-of-consciousness essays with associated Big Five questionnaire results of each respective student.
The researchers focused on the classifier for stability.
The original representation of the numerical factor was simplified to a dichotomy between \emph{positive} and \emph{negative}, denoting essay authors with values in the upper or lower third respectively, and discarding texts from authors in the more ambiguous middle third.
The resulting classifier then achieved an accuracy of 65.7~percent.
Similar performance for the other factors was claimed as well, but not published.


A number of regression models were trained and tested for Big Five analysis on texts in \cite{Bai:2013:PBF:2568488.2568807}.
To obtain training data the authors carried out a personality survey on a microblog site,
which yielded the texts and the personality data from 444 users.
This work is a rare example of the Big Five 
being represented an actual spectrum instead of a dichotomy, 
using an interval $[-1, 1]$.
The performance of the systems was therefore measured
as the deviation from the expected trait values.
The best variant achieved an average \emph{Mean Absolute Percentage Error} (i.e. MAPE over all five traits) of 14~percent.

In \cite{kalghatgi2015neural} the authors used neural networks to analyze the Big Five
personality traits of Twitter users based on their tweets.
The system had no fine-grained scoring, instead classifying each trait only as either \emph{yes} (high) or \emph{no} (low).
The authors did not provide any details about their training data,
and the rudimentary evaluation allows no conclusions regarding the actual performance
of the system.

Deep convolutional neural networks were used in \cite{majumder2017} as classifiers on the Pennebaker \& King dataset of 2,469 Big Five annotated stream-of-consciousness essays \cite{PennebakerKing1999}.
The authors filtered the essays, discarding all sentences
that did not contain any words from a list
of emotionally charged words.
One classifier was then trained for each trait,
with each trait classified only as either \emph{yes} (high) or \emph{no} (low).
The trait classifiers achieved their respective best accuracies using different configurations.
Averaging these best results yielded an
overall best accuracy of 58.83 percent. 

The authors of \cite{TANDERA2017604} trained and evaluated an assortment of Deep Learning networks on two datasets: a subset of the Big Five-annotated \emph{myPersonality} dataset with 10,000 posts from 250 Facebook users,
and another 150 Facebook users whose posts the authors collected manually and had annotated using the \emph{ApplyMagicSauce} tool mentioned above.
The traits were represented in their simplified binary form.
Their best system achieved an average accuracy of 74.17~percent.

In \cite{arnoux201725} the accuracy of works on Big Five personality inference as a function of the size of the input
text was studied. The authors showed that using Word Embedding with Gaussian Processes provided the best results when building a classifier for predicting the personality from tweets. The data consisted of self-reported personality ratings as well as tweets from a set of 1,323 participants. 

In \cite{Yarkoni2010PersonalityI1} a set of 694 blogs
with corresponding self-reported personality ratings was collected. The Linguistic Inquiry and Word Count (LIWC) 2001\footnote{\url{http://liwc.wpengine.com/}} program was used to analyze the blogs. A total of 66 LIWC categories was used for each personality trait.  The results revealed robust correlations between the Big
Five traits and the frequency with which bloggers used different word categories.

\section{Model Training}\label{sec:model_training}

\begin{figure*}[ht!]
\centering
\includegraphics[width=0.8\textwidth]{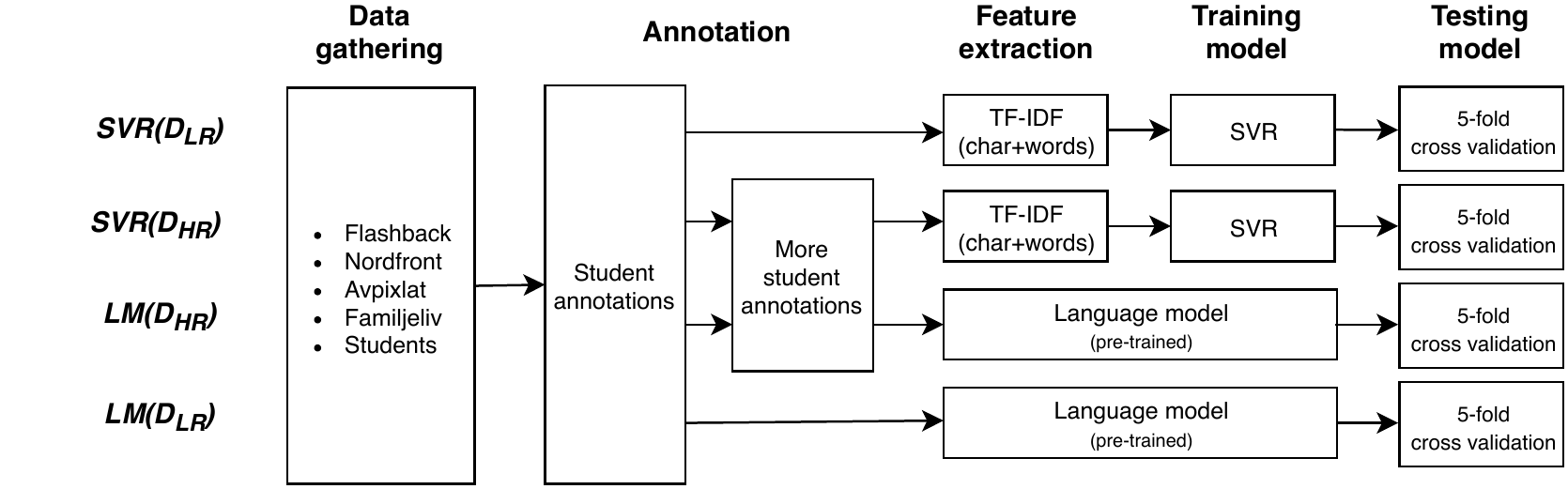}
\caption{Workflow for building the models}
\label{fig:workflow}
\end{figure*}

We employed machine learning for our text-based analysis of the Big Five personality traits.
Applying machine learning presupposes large sets of annotated training data, and our case is no exception.
Since we are working with Swedish language, we could not fall back on any existing large datasets like the ones available for more widespread languages such as English.
Instead our work presented here encompassed the full process from the initial gathering of data over data annotation and feature extraction to training and testing of the detection models.
To get an overview of the process, the workflow is shown in Figure~\ref{fig:workflow}.


Data annotation is time intensive work.
Nevertheless, we decided to assemble two datasets, one prioritizing quantity over quality and one vice versa. The two sets are:
\begin{itemize}
    \item \dlr: a large dataset with \underline{l}ower \underline{r}eliability (most text samples annotated by a single annotator),
    \item \dhr: a smaller dataset with \underline{h}igher \underline{r}eliability (each text sample annotated by multiple annotators).
\end{itemize}

By evaluating both directions we hoped to gain insights into the best allocation of annotation resources for future work. Regarding the choice of machine learning methods we also decided to test two approaches:
\begin{itemize}
    \item support vector regression (SVR): a well-understood method for the prediction of continuous values,
    \item pre-trained language model (LM) with transfer learning: an LM is first trained on large amounts of non-annotated text, learning the relations between words of a given language; it is then fine-tuned for classification with annotated samples, utilizing its language representation to learn better classification with less training data. LM methods currently dominate the state-of-the-art in NLP\footnote{natural language processing} tasks \cite{devlin-etal-2019-bert}.
\end{itemize}


Each method was used to train a model on each dataset, resulting in a total of four models: \svrdlr and \lmdlr denoting the SVR and the language model trained on the larger dataset, and \svrdhr and \lmdhr based on the smaller set with more reliable annotations.



Technically, each of these four models consists of five subvariants,
one for each Big Five personality trait,
though for the sake of simplicity we will keep referring to the four main models only.
Furthermore, to enhance legibility we will omit the dataset denotation in the model name when it is clear from the context which version is meant (e.g. in result tables).

\subsection{Data}\label{sec:data}

As we intended our models to predict the Big Five personality traits on a scale from -3 to 3,
rather than binary classification,
we required training data that contained samples representing the whole data range for each trait.
Given that no such dataset was available for the Swedish language,
we set up our own large-scale collection and annotation operation.

The data was retrieved from four different Swedish discussion forums and news sites.
These sources were selected such as to increase the chances of finding texts from authors with a variety of different personalities.
Specifically, the four sources are:
\begin{itemize}
    \item\textit{Avpixlat}\footnote{\url{https://avpixlat.info} - The site ceased to operate in its original form some time after our data collection, and the address now redirects to a successor with slightly different focus and content.}: a migration critical news site with an extensive comment section for each editorial article. The debate climate in the comment section commonly expresses disappointment towards the society, immigrants, minority groups and the government.
    \item\textit{Familjeliv}\footnote{
    \url{https://www.familjeliv.se}}: a discussion forum with the main focus on family life, relationships, pregnancy, children etc.
    \item\textit{Flashback}\footnote{\url{https://www.flashback.org}}: an online forum with the tagline \emph{``freedom of speech - for real''}, and in 2018 the fourth most visited social media in Sweden\cite{davidsson2018svenskarnaONLINE}. The discussions on Flashback cover virtually any topic, from computer and relationship problems to sex, drugs and ongoing crimes.
    \item\textit{Nordfront}\footnote{\url{https://www.nordfront.se}}: the Swedish news site of the Nordic Resistance Movement (NMR - \emph{Nordiska motst\r{a}ndsr\"{o}relsen}). NMR is a nordic national socialist party. The site features editorial articles, each with a section for reader comments.
\end{itemize}    

Web spiders were used to download the texts from these sources.
In total this process yielded over 70 million texts, but due to time constraints only a small fraction could be annotated and thus form our training datasets \dlr and \dhr.
Table~\ref{tab:numberofsamplespersource} details the size of the datasets, and how many annotated texts from each source contributed to each dataset.
\dhr also contains 59 additional student texts created by the annotators themselves,
an option offered to them during the annotation process (described in the following section). 

\begin{table}[ht]
\centering
\caption{\label{tab:numberofsamplespersource}Number of training samples from each of the data sources per dataset}
\begin{tabular}{lccc}
                       \textbf{Source}    & \multicolumn{1}{c}{\dlr} & \multicolumn{1}{c}{\dhr} & \\
                      \hline
Avpixlat          &  13,064  &  1,100 \\
Familjeliv &      16,322      & 938 \\
Flashback      &  6,324 &305 \\
Nordfront      &   3,660    &  370 \\
Students       &  0  & 59 \\\hline
\textbf{Total} & 39,370 & 2,772 
\end{tabular}
\end{table}

\subsection{\label{sec:annotation} Annotation}
The texts were annotated by 18 psychology students, each of whom had studied at least 15 credits of personality psychology.
The annotation was carried out using a web-based tool.
A student working with this tool would be shown a text randomly picked from one of the sources,
as well as instructions to annotate one of the Big Five traits by selecting a number from the discrete integer interval -3 to 3. 
Initially the students were allowed to choose which of the five traits to annotate,
but at times they would be instructed to annotate a specific trait,
to ensure a more even distribution of annotations.
The tool kept the samples at a sufficiently meaningful yet comfortable size by picking only texts with at least two sentences, and truncating them if they exceeded five sentences or 160 words.

\begin{figure}[ht]
\centering
\includegraphics[width=0.51\textwidth]{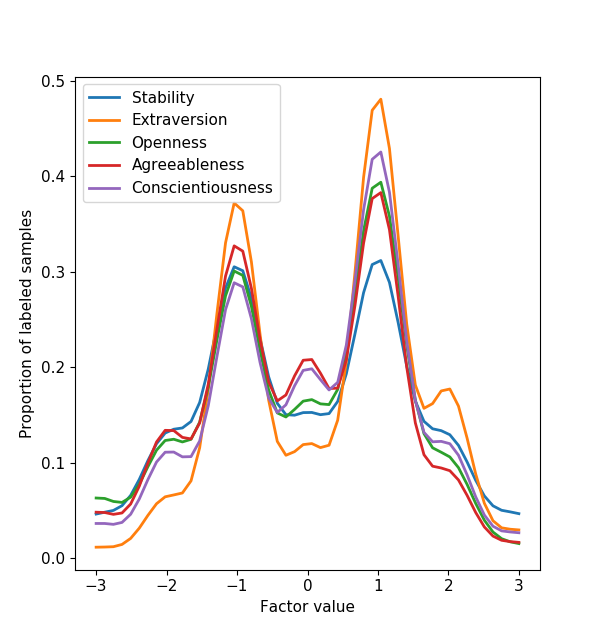}
\caption{\label{fig:dist_large}Distribution of labeled samples for each of the factors of the large dataset.}
\end{figure}

The large dataset \dlr was produced in this manner, with 39,370 annotated texts.
Due to the random text selection for each annotator, the average sample received 1.02 annotations - i.e. almost every sample was annotated by only one student and for only one Big Five trait.
The distribution of annotations for the different factors is shown in Figure~\ref{fig:dist_large}.
We considered the notable prevalence of -1 and 1 to be symptomatic of a potential problem: 
random short texts like in our experiment, often without context, are likely not to contain any definitive personality related hints at all,
and thus we would have expected results closer to a normal distribution.
The students preferring -1 and 1 over the neutral zero might have been influenced by their desire to glean some psychological interpretation even from unsuitable texts.  

For \dhr, the smaller set with higher annotation reliability, we therefore modified the process.
Texts were now randomly selected from the subset of \dlr containing texts which had been annotated with -3 or 3.
We reasoned that these annotations at the ends of the spectrum were indicative
of texts where the authors had expressed their personalities more clearly.
Thus this subset would be easier to annotate,
and each text was potentially more suitable for the annotation of multiple factors. 

\begin{figure}[ht]
\centering
\includegraphics[width=0.51\textwidth]{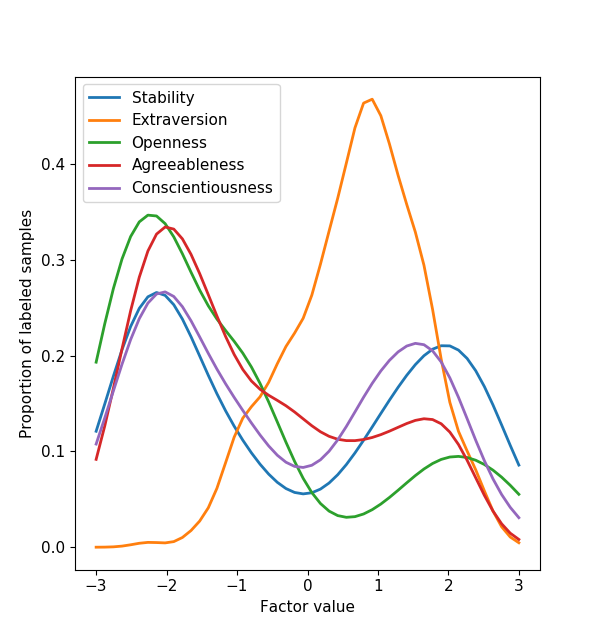}
\caption{Distribution of labeled samples for each of the factors of the small dataset}
\label{fig:dist_small}
\end{figure}

Eventually this process resulted in 2,774 texts with on average 4.5 annotations each.
The distribution for the different factors is shown in Table~\ref{fig:dist_small},
where multiple annotations
of the same factor for one text were compounded into a single average value.



The intra-annotator reliability of both datasets \dlr and \dhr is shown in Table \ref{tab:kripp}.
The reliability was calculated using the Krippendorff's alpha coefficient. Krippendorff's alpha can handle missing values, which in this case was necessary since many of the texts were annotated by only a few annotators. 

\begin{table}[ht]
\caption{\label{tab:kripp}Krippendorff's alpha coefficient for \dlr}

\centering
\begin{tabular}{lc}

\textbf{Factor}          & \textbf{\dlr}  \\ \hline
Stability       & -0.26                               \\
Extraversion      & 0.07                               \\
Openness          & 0.36                              \\
Agreeableness     & 0.51                              \\
Conscientiousness & 0.31                              

\end{tabular}
\end{table}
\begin{table}[ht] 
\centering
\caption{\label{tab:numberofsamplesperfactor}Number of training samples for each of the personality factors}
\begin{tabular}{lccc}
\textbf{Factor}    & \multicolumn{1}{c}{\textbf{\dlr}} & \multicolumn{1}{c}{\textbf{\dhr}} \\\hline
Stability &   8,878  & 817 \\
Extraversion & 7,667  & 358 \\
Openness      &   7,926  & 602 \\
Agreeableness      & 8,564  &   725 \\
Conscientiousness       &   6,877  &    453 \\\hline
\textbf{Total} & 39,912 & 2,774 
\end{tabular}
\end{table}

\begin{figure}[ht]
\centering
\includegraphics[width=0.51\textwidth]{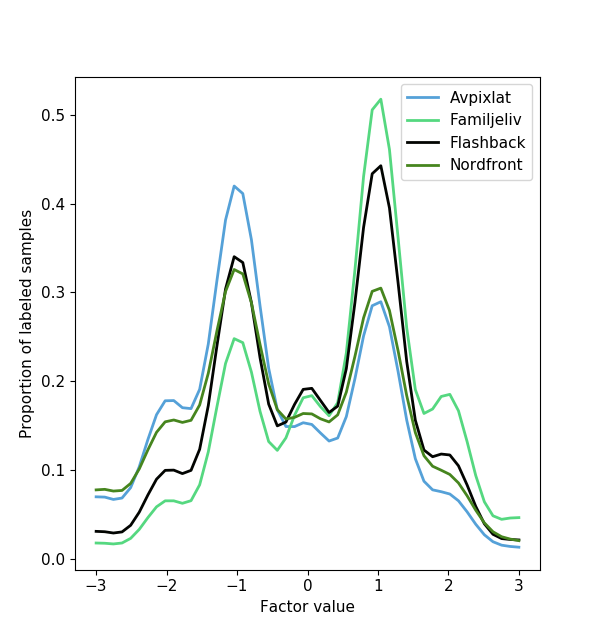}
\caption{\label{fig:dist_sources}Distribution of Big Five factor values for the different sources from the large dataset}
\end{figure}

Table~\ref{tab:numberofsamplesperfactor} shows how many texts were annotated for each factor, and 
Figure~\ref{fig:dist_sources} shows how the different sources span over the factor values.

 Avpixlat and Nordfront have a larger proportion of annotated texts with factors below zero, while Flashback and especially Familjeliv have a larger proportion in the positive interval. The annotators had no information about the source of the data while they were annotating.

\subsection{Feature Extraction}
To extract information from the annotated text data and make it manageable for the regression algorithm, we used Term Frequency-Inverse Document Frequency (TF-IDF) to construct features from our labeled data. TF-IDF is a measurement of the importance of continuous series of words or characters (so called \textit{n-grams}) in a document, where n-grams appearing more often in documents are weighted as less important. TF-IDF is further explained in \cite{tfidf}. In this paper, TF-IDF was used on both word and character level with bi-gram for words and quad-grams for characters.

\subsection{Regression Model}
Several regression models were tested from the \emph{scikit-learn} framework \cite{scikit-learn}, such as RandomForestRegressor, LinearSVR, and KNeighborsRegressor. The Support Vector Machine Regression yielded the lowest MAE and MSE while performing a cross validated grid search for all the models and a range of hyperparameters.    

\subsection{Language Model}
As our language model we used \emph{ULMFiT} \cite{howard2018universal}. ULMFiT is an NLP transfer learning algorithm that we picked due to its straightforward implementation in the \emph{fast.ai} library,\footnote{\url{https://www.fast.ai}} and its promising results on small datasets. As the basis of our ULMFiT model we built a Swedish language model on a large corpus of Swedish text retrieved from the Swedish Wikipedia\footnote{\url{https://sv.wikipedia.org}}
and the aforementioned forums Flashback and Familjeliv.
We then used our annotated samples to fine-tune the language model, resulting in a classifier for the Big Five factors.   

\subsection{Model Performance}
The performance of the models was evaluated with cross validation measuring MAE, MSE and $\textrm{R}^2$.
We also introduced a \textit{dummy regressor}. The dummy regressor is trained to always predict the mean value of the training data. In this way it was possible to see whether the trained models predicted better than just always guessing the mean value of the test data. To calculate the $\text{R}^2$ score we use the following measurement:
$$ \textrm{R}^2 = 1 - \frac{\sum\limits_{i} e^2_i}{\sum\limits_{i} (y_i - \bar{y}_i)^2} $$
where $y$ is the actual annotated score, $\bar{y}$ is the sample mean, and $e$ is the residual.

\subsubsection{Cross Validation Test}
The models were evaluated using 5-fold cross validation. The results for the cross validation is shown in table \ref{tab:cvsmall} and \ref{tab:cvlarge}.


\begin{table*}[t]
\centering
\caption{\label{tab:cvsmall}Cross validated model performance of models trained on  \dhr}
\begin{tabular}{llcc|lcc|lcc}
\textbf{Factor} & $\textrm{MAE}^{}_\textrm{\textit{LM}}$   & $\textrm{MAE}^{}_\textrm{\textit{SVR}}$  & $\textrm{MAE}^{}_\textrm{\textit{Dum}}$ & \multicolumn{1}{l}{$\textrm{MSE}^{}_\textrm{\textit{LM}}$} & $\textrm{MSE}^{}_\textrm{\textit{SVR}}$  & \multicolumn{1}{c|}{$\textrm{MSE}^{}_\textrm{\textit{Dum}}$} & \multicolumn{1}{l}{$\textrm{R}^2_\textrm{\textit{LM}}$} & $\textrm{R}^2_\textrm{\textit{SVR}}$ & $\textrm{R}^2_\textrm{\textit{Dum}}$\\[0.1cm] \hline
Stability         & \textbf{1.01} & 1.23         & 1.84           & \textbf{1.72}                    & 2.24         & 3.85           & \textbf{0.55}                                     & 0.42                          & 0.00                              \\
Extraversion      & \textbf{0.56} & 0.60          & 0.72           & \textbf{0.53}                    & 0.59         & 0.80            & \textbf{0.34}                                     & 0.27                          & 0.00                          \\
Openness          & \textbf{1.01} & 1.09         & 1.37           & \textbf{1.74}                    & 2.11         & 2.98           & \textbf{0.40}                                      & 0.27                          & -0.02                           \\
Agreeableness      & \textbf{0.58} & 0.75         & 1.31           & \textbf{0.57}                    & 0.91         & 2.33           &\textbf{0.75}                                               & 0.61                & -0.01                         \\
Conscientiousness & \textbf{1.06} & 1.28         & 1.61           & \textbf{1.84}                    & 2.24         & 3.08           & \textbf{0.39}                                     & 0.26                          & -0.01  \\\hline
\textbf{Average} & 0.84 & 0.99 & 1.37 & 1.28 & 1.62 & 2.61  & 0.49 & 0.37 & -0.01 \\
\textbf{Total}   & 4.22 & 4.95 & 6.85 & 6.40  & 8.09 & 13.04 & 2.43 & 1.83 & -0.04
\end{tabular}
\end{table*}

\begin{table*}[t]
\centering
\caption{\label{tab:cvlarge}Cross validated model performance of models trained on \dlr}
    \begin{tabular}{llcc|lcc|lcc}
\textbf{Factor} & $\textrm{MAE}^{}_\textrm{\textit{LM}}$   & $\textrm{MAE}^{}_\textrm{\textit{SVR}}$  & $\textrm{MAE}^{}_\textrm{\textit{Dum}}$ & \multicolumn{1}{l}{$\textrm{MSE}^{}_\textrm{\textit{LM}}$} & $\textrm{MSE}^{}_\textrm{\textit{SVR}}$  & \multicolumn{1}{c|}{$\textrm{MSE}^{}_\textrm{\textit{Dum}}$} & \multicolumn{1}{l}{$\textrm{R}^2_\textrm{\textit{LM}}$} & $\textrm{R}^2_\textrm{\textit{SVR}}$ & $\textrm{R}^2_\textrm{\textit{Dum}}$\\[0.1cm] \hline
Stability         & \textbf{1.19} & \textbf{1.19} & 1.26           & 2.15                             & \textbf{2.12} & \multicolumn{1}{c|}{2.20}           & 0.02                                              & \textbf{0.04}                 & 0.00                               \\
Extraversion      & 1.10          & \textbf{1.09} & 1.13           & 1.79                             & 1.72          & \multicolumn{1}{c|}{\textbf{1.64}}  & -0.10                                             & -0.05                         & \textbf{0.00}                      \\
Openness          & 1.13          & \textbf{1.11} & 1.20           & 1.98                             & \textbf{1.89} & \multicolumn{1}{c|}{1.96}           & -0.01                                             & \textbf{0.03}                 & 0.00                               \\
Agreeableness      & \textbf{1.04} & 1.05          & 1.14           & 1.71                             & \textbf{1.70} & 1.79                                &  \ 0.04                                              & \textbf{0.05}                 & 0.00                               \\
Conscientiousness & 1.10          & \textbf{1.09} & 1.13           & 1.90                             & 1.82          & \multicolumn{1}{c|}{\textbf{1.80}}  & -0.06                                             & -0.01                         & \textbf{0.00}   \\\hline
\textbf{Average} & 1.11 & 1.11 & 1.17 & 1.91 & 1.85 & 1.88 & -0.02 & 0.01 & 0.00  \\
\textbf{Total}   & 5.55 & 5.52 & 5.86 & 9.54 & 9.25 & 9.39 & -0.10 & 0.06 & -0.01
\end{tabular}
\end{table*}

For both datasets \dlr and \dhr, the trained models predict the Big Five traits better than the dummy regressor. This means that the trained models were able to catch signals of personality from the annotated data. Extraversion and agreeableness were easiest to estimate. The smallest differences in MAE between the trained models and the dummy regressor are for extraversion and conscientiousness, for models trained on the lower reliability dataset \dlr. The explanation for this might be that both of the factors are quite complicated to detect in texts and therefore hard to annotate. For the models based on \dhr, we can find a large difference between the MAE for both stability and agreeableness. Agreeableness measures for example how kind and sympathetic a person is, which appears much more naturally in text compared to extraversion and conscientiousness. Stability, in particular low stability, can be displayed in writing as expressions of emotions like anger or fear, and these are often easy to identify.

\subsubsection{Binary Classification Test}
As set out in Section~\ref{sec:related_work},
earlier attempts at automatic analysis of the Big Five traits
have often avoided modelling the factors on a spectrum,
instead opting to simplify the task to a binary classification
of \emph{high} or \emph{low}.
We consider our $[-3, 3]$ interval-based representation to be preferable, as it is sufficiently granular to express realistic nuances while remaining simple enough not to overtax annotators with too many choices.
Nevertheless, to gain some understanding of how our approach
would compare to the state of the art,
we modified our methods to train binary classifiers
on the large and small datasets.
For the purposes of this training a factor value below zero
was regarded as \emph{low} and values above as \emph{high},
and the classifiers learnt to distinguish only these two classes.
The accuracy during cross validation was calculated and is presented in Table~\ref{tab:binclas}.
Note that a direct comparison with earlier systems is problematic due to the differences in datasets.
This test merely serves to ensure that our approach
is not out of line with the general performance in the field.


\begin{table*}[t]
\centering
\caption{\label{tab:binclas}Cross validated accuracy for all the binary models on \dhr and \dlr}
    \begin{tabular}{llll|lll}
 & \multicolumn{3}{c|}{\dhr}& \multicolumn{3}{c}{\dlr}\\
\hline
\textbf{Factor} & $\textrm{ACC}^{ }_\textrm{\textit{LM}}$ & $\textrm{ACC}^{ }_\textrm{\textit{SVR}}$  & $\textrm{ACC}^{ }_\textrm{\textit{Dum}}$ & \multicolumn{1}{l}{$\textrm{ACC}^{ }_\textrm{\textit{LM}}$} & $\textrm{ACC}^{ }_\textrm{\textit{SVR}}$ & \multicolumn{1}{c}{$\textrm{ACC}^{ }_\textrm{\textit{Dum}}$} \\[0.1cm] \hline
Stability           & \textbf{0.85}    &     0.80    &    0.48    &  \textbf{0.62}          &  0.59      &  0.51   \\
Extraversion        & \textbf{0.78}    &     0.74    &    0.63    &  \textbf{0.57}           &  0.56      &  0.51   \\
Openness            & \textbf{0.83}    &     0.81    &    0.68    &  \textbf{0.62}          &  0.59      &  0.50   \\
Agreeableness       & \textbf{0.90}    &     0.86    &    0.57    &  \textbf{0.64}          &  0.62      &  0.52   \\
Conscientiousness   & \textbf{0.76}    &     0.71    &    0.52    &  \textbf{0.6}          &  0.56      &  0.51   \\ \hline
\textbf{Average}    & 0.82            &     0.78   &    0.58   &  0.61                   &  0.58     &  0.51   \\
\textbf{Total}      & 4.10             &     3.92    &    2.88    &  3.04                   &  2.92      &  2.55    
\end{tabular}
\end{table*}

We conducted a head-to-head test (paired sample t-test) to compare the trained language model against the corresponding dummy regressor and found that the mean absolute error was significantly lower for the language model \lmdhr, t(4) = 4.32, p = .02, as well as the \lmdlr, t(4) = 4.47, p = .02. Thus, the trained language models performed significantly better than a dummy regressor. In light of these differences and the slightly lower mean absolute error \lmdhr compared to the \lmdlr [t(4) = 2.73, p = .05] and considering that \lmdhr is the best model in terms of $\textrm{R}^2$ we take it for testing in the wild.

\section{Personality Detection in the Wild}\label{sec:wild}
Textual domain differences may affect the performance of a trained model more than expected.
In the literature systems are often only evaluated on texts from their training domain.
However, in our experience this is insufficient to assess
the fragility of a system towards the data,
and thus its limitations with respect to an actual application and generalizability across different domains.
It is critical to go beyond an evaluation of trained models on the initial training data domain,
and to test the systems ``in the wild'',
on texts coming from other sources, possibly written with a different purpose.
Most of the texts in our training data have a conversational nature,
given their origin in online forums,
or occasionally in opinionated editorial articles.
Ideally a Big Five classifier should be able to measure personality traits in any human-authored text
of a reasonable length.
In practice though it seems likely that the subtleties involved in personality detection
could be severely affected by superficial differences in language and form.
To gain some understanding on how our method would perform outside the training domain,
we selected our best model \lmdhr and evaluated it on texts from two other domains.


\subsection{Cover Letters Dataset}
The cover letters dataset was created during a master thesis project at Uppsala University. The aim of the thesis project was to investigate the relationship between self-reported personality and personality traits extracted from texts. In the course of the thesis, 200 study participants each wrote a cover letter and answered a personality form \cite{JOHNSON201478}.  186 of the participants had complete answers and therefore the final dataset contained 186 texts and the associated Big Five personality scores.

\begin{figure}[t]
\centering
\includegraphics[width=0.51\textwidth]{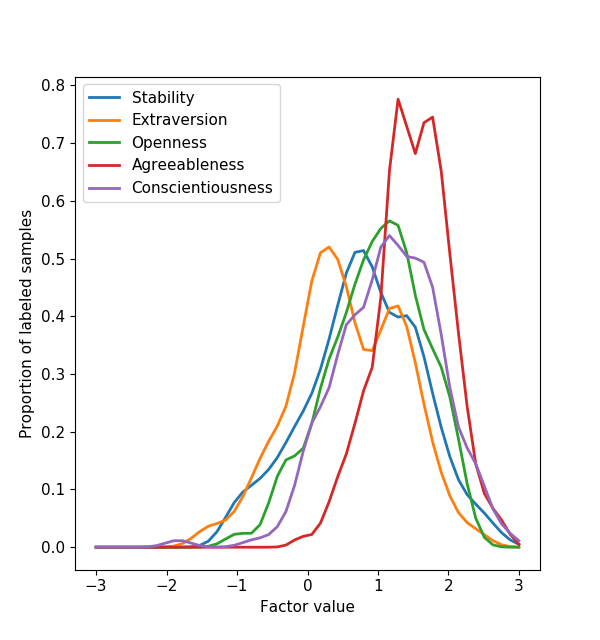}
\caption{\label{fig:dist_test}Distribution of labeled samples for each of the factors of the cover letters dataset}
\end{figure}

We applied \lmdhr to the cover letters to produce Big Five trait analyses, 
and we compared the results to the scores from the personality questionnaire.
This comparison, measured in the form of the evaluation metrics MAE, MSE and $\textrm{R}^2$, is shown in Table~\ref{tab:small_on_ref}. As it can be seen in the table, model performance is poor and $\textrm{R}^2$ was not above zero for any of the factors.

\begin{table}[ht]
\centering
\caption{\label{tab:small_on_ref}Performance of the language model \lmdhr tested on cover letters}
    \begin{tabular}{lc|c|c}
\textbf{Factor}           & MAE & MSE & $\textrm{R}^2$ \\ \hline
Stability         & 0.75 & 0.88 & -0.36 \\
Extraversion      & 0.74           & 0.85    & -0.41           \\
Openness          & 0.74 &  0.8 & -0.69 \\
Agreeableness      & 1.02 & 1.44 & -4.56 \\
Conscientiousness & 0.77        & 0.88       & -0.7      \\\hline
\textbf{Average} & 0.80 & 0.97 & -1.34   \\
\textbf{Total}   & 4.02 & 4.85 & -6.72                        
\end{tabular}
\end{table}

\begin{table}[t]
\centering
\caption{\label{tab:small_on_selftexts}Performance of the language model \lmdhr tested on self-descriptions}
    \begin{tabular}{lc|c|c}
\textbf{Factor}           & MAE & MSE & $\textrm{R}^2$ \\ \hline

Stability &  1.41 &  2.69 &  -2.75  \\ 
Extraversion &  0.84  & 1.15  & -0.89  \\ 
Openness &  0.92 & 1.26  & -1.67 \\ 
Agreeableness & 1.16 & 1.87  & -5.72  \\ 
Conscientiousness &  0.8 &  0.96  & -1.16  \\\hline 
\textbf{Average} & 1.03 &  1.59  & -2.44  \\ 
\textbf{Total} & 5.13 &  7.94  & -12.18 \\ 

\end{tabular}
\end{table}

\subsection{Self-Descriptions Dataset}

\begin{figure}[t]
\centering
\includegraphics[width=0.51\textwidth]{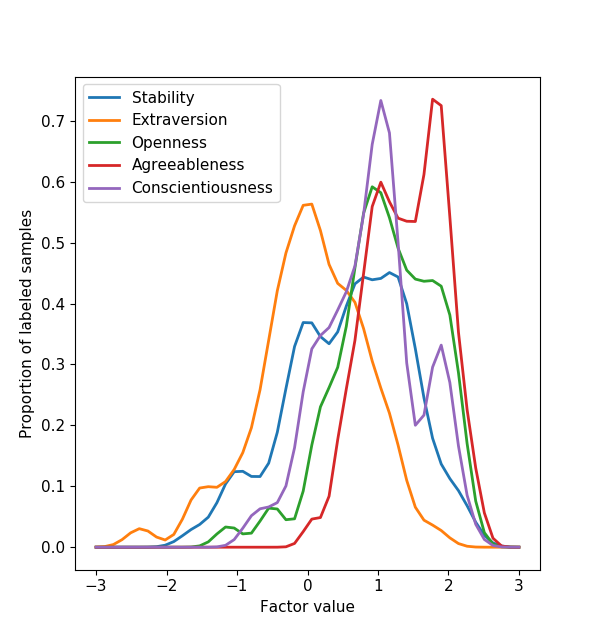}
\caption{\label{fig:dist_test2}Distribution of labeled samples for each of the factors of the self-descriptions letters dataset}
\end{figure}

The self-descriptions dataset is the result of an earlier study conducted at Uppsala University.
The participants, 68 psychology students (on average 7.7 semester), were instructed to describe themselves in text, yielding 68 texts with an average of approximately 450 words. The descriptions were made on one (randomly chosen) of nine themes like politics and social issues, film and music, food and drinks, and family and children. Each student also responded to a Big Five personality questionnaires consisting of 120 items. The distribution of the Big Five traits for the dataset is shown in figure \ref{fig:dist_test2}.

Given this data, we applied \lmdhr to the self-description texts to compute the Big Five personality trait values. We then compared the results to the existing survey assessment using the evaluation metrics MAE, MSE and $\textrm{R}^2$, as shown in Table~\ref{tab:small_on_selftexts}. As it can be seen in the table, model performance was poor and $\textrm{R}^2$, like the results for the cover letters dataset, was not above zero for any of the Big Five factors.


\section{Conclusions}
In this paper, we aimed to create a model that is able to extract Big Five personality traits from a text using machine learning techniques. We also aimed to investigate whether a model trained on a large amount of solo-annotated data performs better than a model trained on a smaller amount of high-quality data. Finally, we aimed to measure model performance in the wild, on data from two domains that differ from the training data. The results of our experiments showed that we were able to create models with reasonable performance (compared to a dummy classifier). These models exhibit a mean absolute error and accuracy in line with state-or-the-art models presented in previous research, with the caveat that comparisons over different datasets are fraught with difficulties. We also found that using a smaller amount of high-quality training data with multi-annotator assessments resulted in models that outperformed models based on a large amount of solo-annotated data. Finally, testing our best model (\lmdhr) in the wild and found that the model could not, reliably, extract people’s personality from their text. These findings reveal the importance of the quality of the data, but most importantly, the necessity of examining models in the wild. Taken together, our results show that extracting personality traits from a text remains a challenge and that no firm conclusions can be made on model performance before testing in the wild. We hope that the findings will be guiding for future research.

\bibliographystyle{IEEEtran}
\bibliography{literature}

\end{document}